\documentclass{article}
\usepackage{ijcai25}

\usepackage{times}
\usepackage{verbatim}
\usepackage{soul}
\usepackage{url}
\usepackage{booktabs}
\usepackage{multirow}
\usepackage[hidelinks]{hyperref}
\usepackage[utf8]{inputenc}
\usepackage[small]{caption}
\usepackage{amsthm}
\usepackage{bm}
\usepackage{balance}
\usepackage{booktabs}
\usepackage[switch]{lineno}

\usepackage{subfig}
\usepackage{graphicx}
\usepackage[utf8]{inputenc}
\usepackage{array}
\usepackage{nccmath}
\usepackage{amsmath,bm}
\usepackage{amssymb}
\usepackage{mathrsfs} 
\usepackage{diagbox}
\usepackage{mathtools, nccmath, relsize}
\usepackage{enumitem}
\usepackage{pbox}
\usepackage{longtable}
\usepackage{chngpage}
\usepackage{amsfonts}
\usepackage{makecell, multirow, tabularx}
\usepackage{algorithm}
\usepackage{algpseudocode}
\usepackage{graphics}

\newcolumntype{Y}{>{\centering\arraybackslash}X}
\usepackage{threeparttable}

\newcommand{\overbar}[1]{\mkern 1.5mu\overline{\mkern-1.5mu#1\mkern-1.5mu}\mkern 1.5mu}

\PassOptionsToPackage{table,xcdraw}{xcolor}

\usepackage{makecell}
\usepackage{colortbl}

\usepackage{multirow}

\usepackage{calc}
\newcolumntype{C}[1]{>{\centering\arraybackslash}p{#1}}%
\newlength{\mycolwidth}
\newlength{\mycolwidthadd}
\newlength{\mytwocolwidthadd}
\newlength{\myfourcolwidth}
\newlength{\mythreecolwidth}
\newlength{\mytwocolwidth}
\usepackage{pifont}

\title{Early Detection of Patient Deterioration from Real-Time Wearable Monitoring System}

\author{
Lo Pang-Yun Ting$^1$
\and
Hong-Pei Chen$^1$
\and
An-Shan Liu$^1$
\and
Chun-Yin Yeh$^{2}$
\and
\\
Po-Lin Chen$^{3}$
\And
Kun-Ta Chuang$^1$\\
\affiliations
$^1$Dept. of Computer Science and Information Engineering, National Cheng Kung University, Taiwan\\
$^2$Dept. of Nursing, National Cheng Kung University, Taiwan\\
$^3$Division of Infectious Diseases, National Cheng Kung University Hospital, Taiwan\\
\emails
\{lpyting, hpchen, asliu\}@netdb.csie.ncku.edu.tw,
z10809062@email.ncku.edu.tw,
\\
\{cplin, ktchuang\}@mail.ncku.edu.tw
}

\begin{document}
\maketitle

\begin{abstract}

Early detection of patient deterioration is crucial for reducing mortality rates. Heart rate data has shown promise in assessing patient health, and wearable devices offer a cost-effective solution for real-time monitoring. However, extracting meaningful insights from diverse heart rate data and handling missing values in wearable device data remain key challenges.  To address these challenges, we propose \emph{TARL}, an innovative approach that models the structural relationships of representative subsequences, known as shapelets, in heart rate time series. \emph{TARL} creates a shapelet-transition knowledge graph to model shapelet dynamics in heart rate time series, indicating illness progression and potential future changes. We further introduce a transition-aware knowledge embedding to reinforce relationships among shapelets and quantify the impact of missing values, enabling the formulation of comprehensive heart rate representations. These representations capture explanatory structures and predict future heart rate trends, aiding early illness detection. We collaborate with physicians and nurses to gather ICU patient heart rate data from wearables and diagnostic metrics to assess illness severity and evaluate deterioration. Experiments on real-world ICU data demonstrate that \emph{TARL} achieves both high reliability and early detection. A case study further showcases \emph{TARL}'s explainable detection process, highlighting its potential as an AI-driven tool to assist clinicians in recognizing early signs of patient deterioration.

\end{abstract}
\section{Introduction}
\label{sec:intro}

Identifying patients at high risk of deterioration is a crucial issue in intensive care units (ICUs). 
ICUs specialize in treating severely ill patients with more resources and a higher nurse-to-patient ratio than general wards. However, ICU patients can deteriorate rapidly, and unnoticed changes may delay critical treatments or transfers. Each hour of delay can increase mortality odds by 3\%~\cite{Churpek2016AssociationBI}, and deaths may occur during such transitions. These challenges highlight the necessity for automated systems that aid clinicians in the early identification of patient deterioration, thereby facilitating prompt interventions and decreasing mortality rates.

Previous research~\cite{Shaffer2017AnOO} has identified patient's heart rate data as a key indicator of inflammation, which is associated with various chronic diseases in adult ICU patients. This implies that consistently monitoring heart rate could be promising for the early identification of patient deterioration. Nonetheless, equipment for ICU monitoring can be expensive and difficult to obtain. Wearable devices provide a cheaper option for ongoing vital sign monitoring, particularly in settings where resources are limited. Grand View Research~\cite{GrandViewResearch} valued the wearable technology market at USD 61.30 billion in 2022, emphasizing its growing potential to enhance patient care via real-time monitoring and early intervention.

Meanwhile, ensuring that AI models provide justifiable outcomes is crucial for clinical decision-making. AI models need to deliver justifiable outcomes to help clinicians assess patient conditions and guide trusted treatment decisions~\cite{Tonekaboni2019WhatCW}. In this context, heart rate time series data offers a rich source of information for identifying critical health patterns. A promising method for examining heart rate time series is shapelet-based analysis, which has shown effectiveness in multiple fields~\cite{Zhu2021NetworkedTS,Li2022EfficientSD}. A \textit{Shapelet}~\cite{Ye2009TimeSS} refers to a time-series subsequence recognized as a characteristic waveform designed to offer clear explanatory insights into time series data.

However, two main challenges remain in analyzing patient heart rates to detect patient illness deterioration through wearable devices. \underline{\textbf{(\textit{i})}} First, heart rate variations are complex and differ among patients, making it difficult to effectively capture explanatory insights from shapelets within diverse heart rate time series data. \underline{\textbf{(\textit{ii})}} Second, data collected by wearable devices often have a high rate of missing values~\cite{Hicks2019BestPF}, which can hinder heart rate analysis. Directly discarding incomplete data may also result in significant information loss. Addressing these challenges is essential for improving the reliability and effectiveness of AI-driven systems for the early detection of illness deterioration.

Therefore, to address these challenges, we propose \emph{\textbf{TARL}} (\underline{\textbf{T}}ransition-\underline{\textbf{A}}ware \underline{\textbf{R}}epresentation \underline{\textbf{L}}earning), a novel approach that integrates shapelet-based knowledge representation while accounting for missing values. To tackle the first challenge, we construct a \textbf{shapelet-transition knowledge graph} from patients' heart rate time series collected from wearable devices. This graph captures the occurrence order of shapelets, preserving their sequential patterns to represent patient illness progression and potential future changes. We then introduce a novel \textbf{transition-aware knowledge embedding} to learn shapelet representations, using an \underline{attention mechanism} to reinforce the relative structure of shapelets. Additionally, we incorporate \underline{transition confidence} to quantify the impact of missing values during embedding learning, ensuring robustness against incomplete data. This specifically addresses the second challenge.

Finally, the learned shapelet representations are used to formulate patient heart rate representations, which embed the explanatory structure of shapelets, potential future heart rate changes, and the impact of missing values. These representations enable early detection of deterioration based on observed shapelet transitions in newly received heart rate data, making the detection process explainable.

To ensure that \emph{TARL} can effectively assist clinicians in clinical practice, we collaborate with physicians and nurses at the hospital to collect IC patient heart rate data using wearable devices. Also, we collect patients' laboratory test results and diagnostic metrics to estimate Apache II scores (Acute Physiology and Chronic Health Evaluation II) ~\cite{Knaus1985apacheii}, a widely used indicator of illness severity that serves as ground truth for evaluating patient deterioration in this paper. Our method is assessed through real-world ICU patient data, measuring deterioration detection accuracy and earliness to ensure its reliability and impact on assisting clinicians. The interdisciplinary research effort integrates AI researchers, clinicians, and nursing staff, using low-cost wearables and an effective model to strengthen hospital monitoring infrastructure. It has the potential to address the needs of frontline clinicians, supports faster interventions for ICU patients, and offers evidence hospitals can use to improve their triage process.

\section{Related Works}

\noindent\textbf{Early Detection of Patient Deterioration.} 
Several studies have explored various methodologies for predicting ICU admission, early diagnosis of specific diseases, and mortality risk. These methods include deep learning~\cite{Li2020DeepAlertsDL,Shamout2020DeepIE,shah2021simulated,salehinejad2023contrastive,ashrafi2024deep,boussina2024impact}, ensemble-based methods~\cite{ElRashidy2020IntensiveCU,Alshwaheen2021ANA,saleh2024predicting,saif2024early}, and detection based on early warning scores~\cite{Paganelli2022ANS,kia2020mews++,romero2021using}. Some approaches leverage interpretable patterns, such as shapelets, to extract critical patient information~\cite{zhao2019asynchronous,Hyland2020EarlyPO,onwuchekwa2024enhanced}

\noindent\textbf{Early Time-Series Classification.}
Early time-series classification methods can be categorized into three groups: prefix-based, model-based, and shapelet-based.
Prefix-based methods determine the minimum prefix length (MPL) required for classification by learning from training instances and applying the learned prefix length to test data. Xing \textit{et al.} first introduce MPL and propose ECTS (early classification on time series)~\cite{Xing2012EarlyCO}. Other works focus on estimating the Minimum Required Length (MRL) for early classification~\cite{Gupta2020AnEC,Gupta2021AFE}. Various issues are discussed in ~\cite{wu2021early}.
Model-based methods use mathematical models to balance earliness and accuracy~\cite{Lv2019AnEC,Schfer2019TEASEREA}. Shapelet-based methods extract key shapelets to enable reliable label prediction for incomplete time series~\cite{Yan2020ExtractingDF}.

Although these methods perform well in their respective settings, they often overlook the structural relationships among time-series patterns and how they evolve over time. While some utilize predefined shapelet extraction, they fail to capture shapelet dependencies and transition dynamics, limiting their ability to model temporal progression. This is particularly critical in early detection of patient deterioration, where understanding patients' condition transitions is essential for early and accurate illness deterioration detection.

\section{{Preliminary}}\label{sec:preliminary}

\begin{figure}
\graphicspath{{figs/}}
\begin{center}
\includegraphics[width=0.5\textwidth]{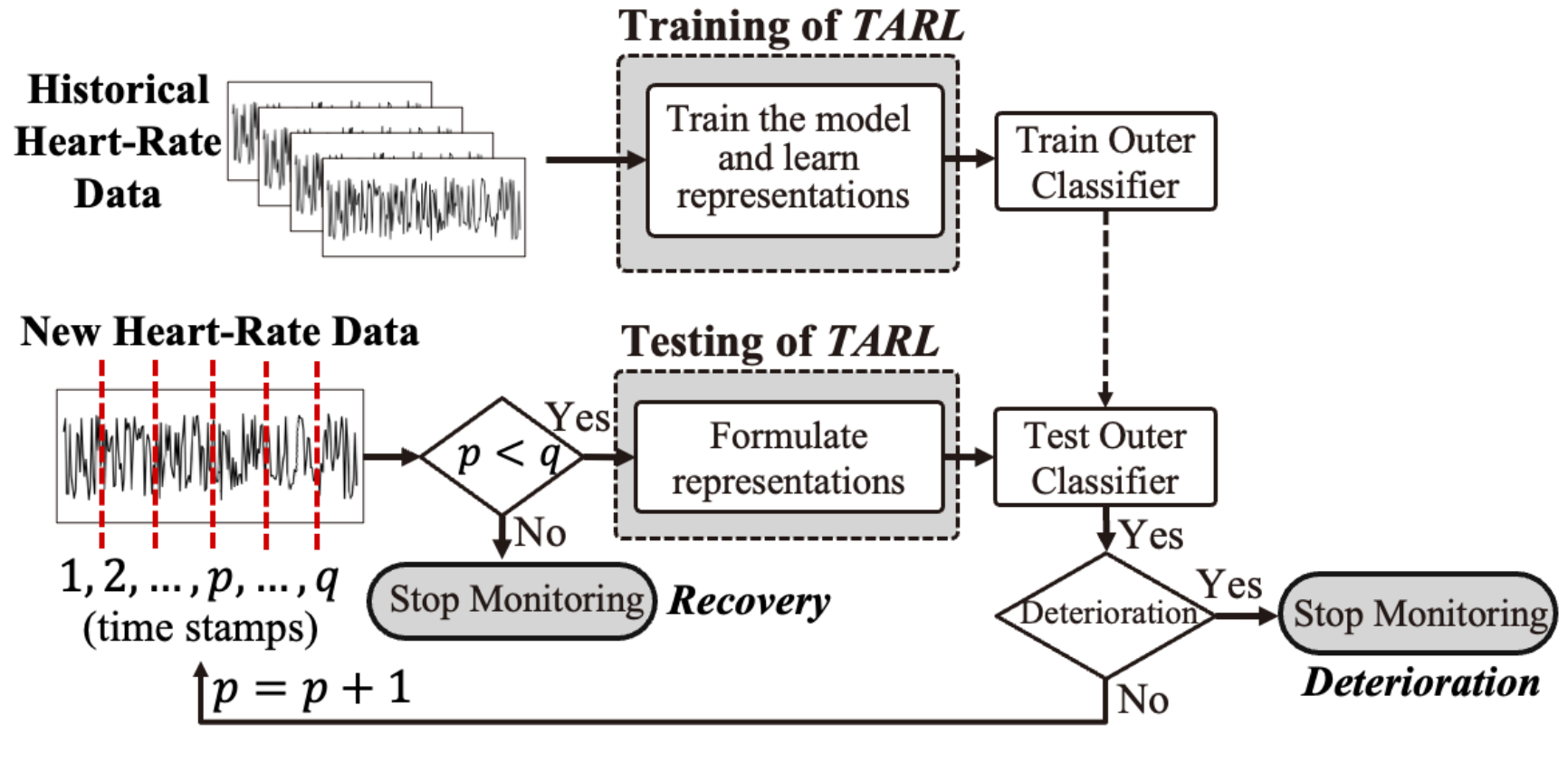}
\end{center}
\caption{The early patient deterioration process with a real-time wearable system.}
\label{fig:scenario}
\end{figure}

\figurename~\ref{fig:scenario} illustrates the proposed real-time wearable monitoring system. During \emph{TARL}’s training, historical heart-rate data from wearable devices are acquired. As monitoring continues, new heart-rate data arrive at fixed intervals until the patient is identified as deteriorating or receives a confirmed diagnosis. \emph{TARL} focuses on developing reliable heart-rate representations to identify early deterioration symptoms, serving as features for early detection classification models.

\noindent\textbf{Definition 1 (Historical Data):} Historical data from all patients is denoted as $\mathcal{X}$, where each $X\in \mathcal{X}$ represents a patient's heart rate time series, expressed as $X=\{x_1, ..., x_{|X|}\}$. Each $x\in\mathbb{R}^+$ is a heart rate data point arranged chronologically. Wearable device data may contain many missing values. To handle missing values in wearable device data, we apply linear interpolation to complete $X$.

\begin{figure*}[t]
\graphicspath{{figs/}}
\begin{center}
\includegraphics[width=0.95\textwidth]{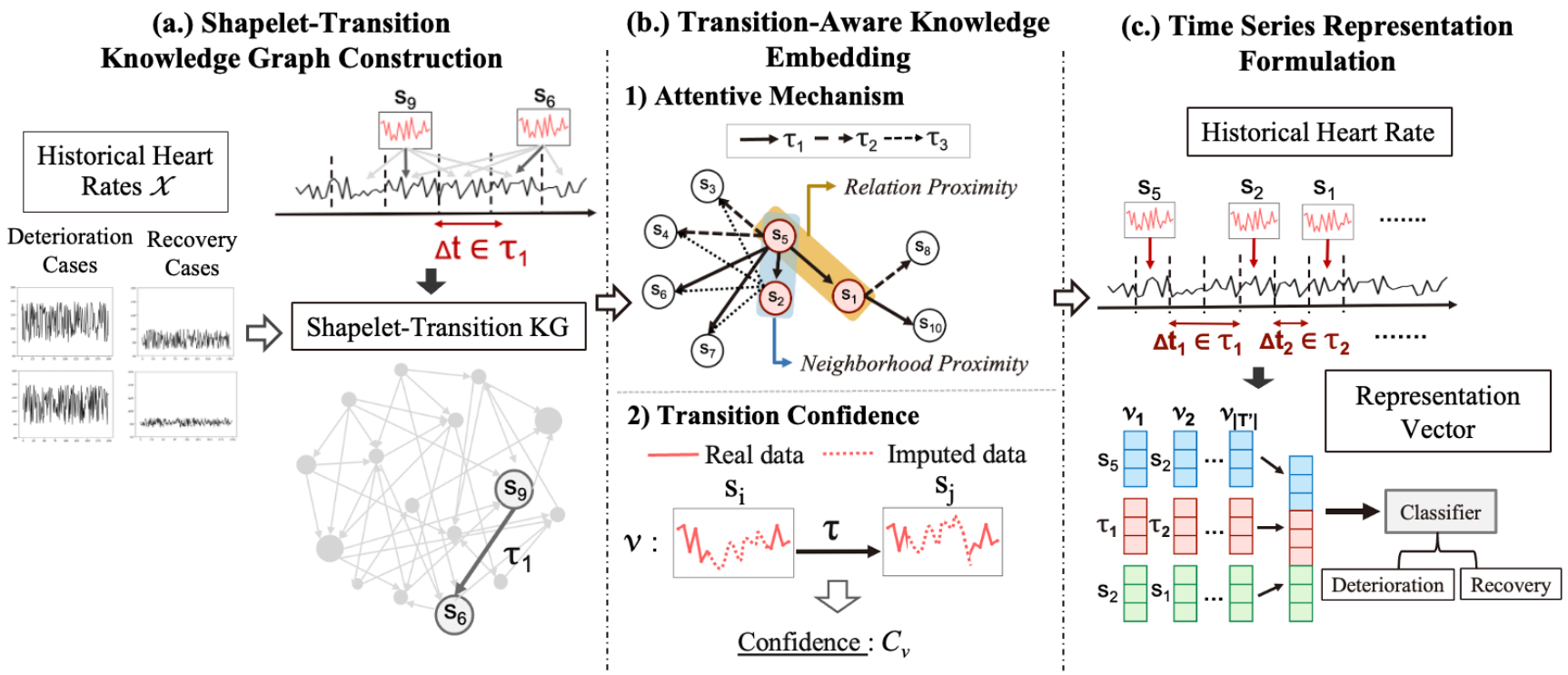}
\end{center}
\caption{The overview of the \emph{TARL} framework. (a) A shapelet-transition knowledge graph is constructed from historical heart rate data. (b) A transition-aware knowledge embedding is introduced to model relationships among shapelets on the graph using an attention mechanism and transition confidence. (c) Heart rate representations are formulated from the embedding results and used as features for classifiers to detect illness deterioration.}
\label{fig:framework}
\end{figure*}

\noindent\textbf{Definition 2 (Deterioration Label):} For each heart rate time series $X \in \mathcal{X}$ measured from a patient, a deterioration label $\hat{l}_X \in \{1, 0\}$ is assigned, where $\hat{l}_X=1$ denotes that the patient’s condition is deteriorating, and $\hat{l}_X=0$ denotes improvement or recovery.

In our setting, deterioration labels (Def. 2) are derived from each patient's Apache II score (Acute Physiology and Chronic Health Evaluation II)~\cite{Knaus1985apacheii}, a severity-of-illness indicator computed from various diagnostic metrics, including age, medical history, respiratory rate, white blood cell count, Glasgow Coma Scale, etc. Following~\cite{Godinjak2016PredictiveVO}, we partition Apache II scores into \underline{four illness levels}: 0–9, 10–19, 20–29, and above 30, with higher levels indicating greater severity and mortality risk. An increase in level indicates deterioration, and a decrease signifies recovery. Thus, if a patient's illness level computed at the time of the last heat rate data point in $X$ exceeds that at the first, the label $\hat{l}_X$ is set to 1; otherwise, it is set to 0.

\noindent\underline{\textbf{Problem Statement:}} At each time step $p$, given historical data $\mathcal{X}$, corresponding deterioration labels $\{\hat{l}_X|X \in \mathcal{X}\}$, and a newly received heart rate time series $X'_{1:p}= \{x_1, x_2, ..., x_p\}$ measured from a patient from time step 1 to $p$, our goal is to detect early whether $X'_{1:p}$ belongs to a deteriorating patient. The detected label is denoted by $l_{X'_{1:p}}\in\{0,1\}$. 
Let the clinicians' diagnosis for $X'$ be determined at time step $q$ (when $\hat{l}_{X'}$ is obtained). A deterioration detection is considered \underline{successful if $l_{X'_{1:p}} = \hat{l}_{X'}=1$} and \underline{$p < q$}. The smaller the time gap between detection and diagnosis, the more timely and effective the detection.
\section{The \emph{TARL} Framework}
\label{sec:framework}

The architecture of \emph{TARL} is illustrated in \figurename~\ref{fig:framework}. It comprises three main components: capturing patient illness progression using a shapelet-transition knowledge graph (Sec.~\ref{subsec:kg}), introducing a transition-aware knowledge embedding (Sec.~\ref{subsec:kge}) to formulate heart rate representations (Sec.~\ref{subsec:representation}), and using these representations as features for early illness deterioration detection.

\subsection{Shapelet-Transition Knowledge Graph Construction}\label{subsec:kg}

To provide a more explainable detection process, we capture explanatory insights from shapelets in patients' heart rate data. Specifically, we seek to understand which shapelets appear during a patient's deterioration process and how they evolve over time. The \underline{occurrence order} and \underline{time intervals} of shapelets in heart rate series are crucial for understanding heart rate dynamics and patient condition progression. We first define key shapelets and their relations as follows.

\noindent\textbf{Definition 3 (Key Shapelet):}  The set of key shapelets is denoted as $S = \{s_1, s_2, ...\}$. Each shapelet $s\in \mathbb{R}^k$ represents $k$ continuous heart rate values obtained from historical heart rate data $\mathcal{X}$ (Def. 1) using the shapelet discovery method~\cite{Grabocka2014LearningTS}.

\noindent\textbf{Definition 4 (Relation):} A relation set is denoted as $\mathcal{T}=\{\tau_1, \tau_2, ...\}$, where each $\tau$ represents a specific range of time intervals between consecutive shapelet occurrences in a heart rate time series. In this work, we define the time interval as 30 minutes, meaning $\tau_1$ corresponds to 0-30 minutes, $\tau_2$ to 30-60 minutes, and so on.

With defined key shapelets and relations, we formulate the \textit{shapelet-transition knowledge graph} to model illness progression in heart rate data.

\noindent\textbf{Definition 5 (Shapelet-Transition Knowledge Graph):} A shapelet-transition knowledge graph (abbreviated as shapelet-transition KG) is defined as $G=\{(s, \tau, s') | s, s' \in S, \tau \in \mathcal{T}\}$, where $S$ is the key shapelet set, and $\mathcal{T}$ is the relation set. Each triplet structure $(s, \tau, s')$ represents a transition from shapelet $s$ to $s'$ in a heart rate time series. 

The example of constructing shapelet-transition KG is shown in \figurename~\ref{fig:framework}a. Shapelets $s_9$ and $s_6$ are identified in a heart rate time series, with a time interval corresponding to relation $\tau_1$. This forms the triplet ($s_9, \tau_1, s_6$), and all such triplets extracted from historical heart rate data collectively form the shapelet-transition KG. To determine whether a shapelet $s\in S$ are identified in a heart rate time series $X\in \mathcal{X}$, we use a matching process: $s$ is considered a match if its the Euclidean distance between to a segment $v$ in $X$ is below a fixed threshold.

\subsection{Transition-Aware Knowledge Embedding}\label{subsec:kge}

We propose a method to embed transition relations among shapelets for heart rate time series representation. To enhance representation reliability, we incorporate an \textbf{attentive mechanism} to reinforce shapelet structure and quantify the impact of missing values through considering \textbf{transition confidence}.

\subsubsection{Attentive Mechanism}
For a target shapelet $n$, its neighbor $m$ should be closer in the embedded space if they share similar structural information, indicating that shapelet $m$ has a higher impact on shapelet $n$. Inspired by the \textit{attention mechanism} commonly used in neural networks to emphasize important features, we design an attentive mechanism to estimate neighbor impact based on the structural equivalence on shapelet-transition KG and enhance representation by aggregating neighbor embeddings. Given a triplet $(s_i, \tau, s_j)$ in the KG $G$, we use neighborhood and relation proximity to enhance representations.

\noindent\underline{\textit{Neighborhood Proximity.}} 
If $s_i$ and $s_j$ are often followed by similar shapelets in $G$, they likely cause similar heart rate variations in patients' time series, so their embeddings should be similar. Formally, let $S_{i, *}$ and $S_{j, *}$ be the outgoing neighbors of $s_i$ and $s_j$, respectively, and let $\overbar{\mathbf{v}}^{t}_{\mathcal{S_{i,*}}}\in \mathbb{R}^d$  and $\overbar{\mathbf{v}}^{t}_{\mathcal{S}_{i,*}}\in \mathbb{R}^d$ be the average embeddings of shapelets in $S_{i, *}$ and $S_{j, *}$ at training epoch $t$, where $d$ is the embedding dimension. The neighborhood proximity between $s_i$ and $s_j$ is:

\begin{equation}\label{eq:neighbor_proximity}
{\small	
\begin{gathered}
\alpha^n_{i,j}=\frac{\text{exp}\left ( e^n_{i,j} \right )}{\sum_{k\in S_{i,*}}^{} \text{exp}\left ( e^n_{i,k} \right )},\\
e^n_{i,j}=jac\Bigl(S_{i,*},S_{j,*}\Bigr)\cdot cos\Bigl(\overbar{\mathbf{v}}^{t}_{\mathcal{S}_{i,*}},\overbar{\mathbf{v}}^{t}_{\mathcal{S}_{j,*}} \Bigr),
\end{gathered}
}
\end{equation}
where the function $jac(S_{i,*},S_{j,*})$ computes the Jaccard similarity between two sets, and  $cos(\overbar{\mathbf{v}}^{t}_{\mathcal{S}_{i,*}},\overbar{\mathbf{v}}^{t}_{\mathcal{S}_{j,*}} )$ measures the cosine similarity of their neighborhood embeddings. $jac(\cdot, \cdot)$ captures the structural similarity of $s_i$ and $s_j$, whereas $cos(\cdot, \cdot)$ reflect their embeddings similarity.

\noindent\underline{\textit{Relation Proximity.}} Unlike neighborhood proximity, relation proximity is determined by outgoing relations. If $s_i$ and $s_j$ typically transition within shorter time intervals, they are likely associated with rapid heart rate changes, and their embeddings should be similar. Given relations sets $\mathcal{T}_{i,*}$ and $ \mathcal{T}_{j,*}$, representing the outgoing relations of $s_i$ and $s_j$, respectively, and $\overbar{\mathbf{v}}^{t}_{\mathcal{T}_{i,*}} \in \mathbb{R}^d$  and $\overbar{\mathbf{v}}^{t}_{\mathcal{T}_{j,*}}\in \mathbb{R}^d$,  denoting the average embeddings of relations in $\mathcal{T}_{(i*)}$ and $\mathcal{T}_{(j*)}$ at the training epoch $t$, the relation proximity between shapelets $s_i$ and $s_j$ is:

\begin{equation}\label{eq:relation_proximity}
{\small	
\begin{gathered}
\alpha^r_{i,j}=\frac{\text{exp}\left ( e^r_{i,j} \right )}{\sum_{k\in S_{i,*}}^{} \text{exp}\left ( e^r_{i,k} \right )},\\
e^r_{i,j}= jac\Bigl(\mathcal{T}_{i,*}, \mathcal{T}_{j,*}\Bigr) \cdot cos\Bigl(\overbar{\mathbf{v}}^{t}_{\mathcal{T}_{i,*}},\overbar{\mathbf{v}}^{t}_{\mathcal{T}_{j,*}} \Bigr),
\end{gathered}
}
\end{equation}
where $jac(\cdot, \cdot)$ and $cos(\cdot, \cdot)$ are as defined in Eq.~\ref{eq:neighbor_proximity}.

Focusing on outgoing neighborhoods and relations reinforces next-shapelet and relation information during training, enabling the capture of early illness symptoms and improving detection earliness. Using Eq.~\ref{eq:neighbor_proximity} and Eq.~\ref{eq:relation_proximity}, the embedding of shapelet $s_i$ at epoch $t$ can be formulated from its outgoing neighbors' embeddings $\mathbf{v}^t_j \in \mathbb{R}^d$, as defined below: 
\begin{equation}
\label{eq:head_update}
{\small	
\mathbf{v}'^t_{i} = \sum_{s_j\in S_{(i,*)}}^{}\Bigl(\omega \alpha^n_{i,j} + (1-\omega)\alpha^r_{i,j}\Bigr)\mathbf{v}_{j}^{t},
}
\end{equation}
where $\omega >0$ is the weight balancing two proximities. With the designed attentive mechanism, shapelet embeddings can be enhanced by prioritizing to more informative neighbors.

\subsubsection{Transition Confidence}  
To ensure robust embedding learning, the confidence of a shapelet’s transition should be considered, which refers to the confidence of a triplet $(s_i, \tau, s_j)$ (transition from $s_i$ to $s_j$ with time interval $\tau$), given the presence of missing values in raw heart rate data collected from wearable devices.

As defined in Def. 1, missing values in wearable device data are imputed using linear interpolation. Subsequently, we estimate the confidence of a data point $x_p$ in the imputed heart rate time series $X$. Let $\Delta t_{p,q}$ be the time difference between $x_p$ and the nearest non-imputed data point $x_q$ in time series $X$. The confidence $\gamma(x_p)$ of data point $x_p$ is defined as:

\begin{equation}
\label{eq:data_re}
\gamma(x_p)=
 \Big(1 - \frac{\Delta t_{p,q}}{\varphi }\Big)\cdot \mathbb{I}(\Delta t_{p,q}< \varphi)
\end{equation}
where $\varphi>0$ is a fixed value, and $\mathbb{I}(\cdot)$ is an indicator function. The smaller the time difference $\Delta t_{p,q}$, the higher the confidence of $x_p$. Note that $\gamma(x_p)$ is set to 1 if $x_p$ is a non-imputed data point. 

We then use Eq.~\ref{eq:data_re} to formulate the transition confidence of a triplet $\nu=(s_i, \tau, s_j)$ in shapelet-transition KG. Assuming shapelets $s_i \in \mathbb{R}^k$ and $s_j\in \mathbb{R}^k$ correspond to segments $v_i = \{x_1, .., x_k\}$ and $v_j = \{x'_1, .., x'_k\}$ of a heart rate time series, respectively, the transition confidence of $\nu$ is formulated as:

\begin{equation}
\label{eq:shapelet_re}
\mathcal{C}({\nu}) = f_{\nu}\cdot\frac{\sum_{x \in v_i}^{ }\gamma(x)\sum_{x \in v_j}^{ }\gamma(x')}{|s_i||s_j|},
\end{equation}
where $f_{\nu}$ represents the occurrence proportion of the transition from $s_i$ to $s_j$ with time interval $\tau$ in the historical data.

\subsubsection{Model Training}

Based on the aggregated embeddings for shapelets (Eq.~\ref{eq:head_update}) and the transition confidence of each triplet (Eq.~\ref{eq:shapelet_re}), we design a triplet-based embedding optimization to reinforce shapelet transition structures. To capture the relationships among shapelets effectively, we employ a margin-based objective. Let $T$ denote the set of all triplets $\nu=(s_i, \tau, s_j)$ in the shapelet-transition KG. The loss function for the transition-aware knowledge embedding is formulated as:

\begin{equation}
\label{eq:loss}
L=\sum_{\nu \in T}^{}\sum_{\bar{\nu} \in \bar{T}}^{}\mathcal{C}_{\nu}\cdot\Bigl[E_{\nu}+\xi -E_{\bar{\nu}}\Bigr]_+,
\end{equation}
where $T'$ represents negative triplets, including all triplet combinations excluding $T$. $\mathcal{C}_{\nu}$ denotes transition confidence (Eq.~\ref{eq:shapelet_re}), and $[\mu]_{+}=\max(0,\mu)$. The margin $\xi > 0$ is used to separate the embeddings of positive and negative triplets. The scoring function $E_{\nu}$ evaluates triplet plausibility, ensuring the structure captures meaningful relationships among shapelets based on the concept of KG embedding. We adopt the DistMult model~\cite{Yang2014EmbeddingEA} as the scoring function:

\begin{equation}
\label{eq:score_func}
E_{\nu} = (\mathbf{v}'^t_i)^{\text{T}}\cdot \text{diag}(\mathbf{v}^t_{\tau})\cdot(\mathbf{v}'^t_j),
\end{equation}
where $\mathbf{v}^t_{\tau}\in \mathbb{R}^d$ is the relation $\tau$'s embedding, and $\mathbf{v}'^t_i$ is the aggregated embedding of shapelet $s_i$ from Eq.~\ref{eq:head_update}. $(\cdot)^{\text{T}}$ denotes transpose, and $\text{diag}(\cdot)$ the diagonal matrix.

By refining the loss function, the embeddings for shapelets and the relations within the shapelet-transition KG can adeptly encapsulate the relationship structures among shapelets, handle missing data, and represent potential changes in heart rate dynamics.

\subsection{Time Series Representation Formulation}\label{subsec:representation}

Utilizing the embeddings derived from the shapelet-transition KG, each heart rate time series $X$ can then be represented to reflect the patient's condition changes.

Let $T_X$ denote the set of triplets occurring in $X$. For each triplet $\nu \in T_X$, let $0<I_{\nu}\leq |T_X|$ represent its \textit{reverse index} in $T_X$, where $I_{\nu}=1$ corresponds to the most recent triplet and $I_{\nu}=|T_X|$ refers to the earliest one. A higher $I_{\nu}$ value indicates an earlier occurrence of $\nu$ in time series $X$. Given the reverse index $I_{\nu}$ for each triplet $\nu \in T_X$, the representation $\bm{\Psi}_X \in \mathbb{R}^{3d}$ of the time series $X$ can be formulated as:

\begin{equation}
\label{eq:timeseries_represent}
\bm{\Psi}_X=\sum_{\nu=(s_i,\tau,s_j)\in T_X}^{} \epsilon^{I_{\nu}}\cdot\frac{\Bigl[\mathbf{v}_i||\mathbf{v}_{\tau}||\mathbf{v}_j\Bigr]}{|T_X|},
\end{equation}
where $\mathbf{v}_i$, $\mathbf{v}_{\tau}$, and $\mathbf{v}_j$ are the learned embeddings of $s_i$, $\tau$, and $s_j$, respectively. The symbol $||$ represents vector concatenation. The fixed value $0<\epsilon<1$ ensures that the most recent triplet in the time series is assigned greater importance.

Consequently, the representations of historical heart rate time series serve as input features for training classification models. As incoming heart rate time series are collected at each time interval, their representations can be generated in a similar manner, allowing for continuous identification of deterioration at every interval.

\section{Experimental Results}

\subsection{Dataset and Experimental Setup}

\subsubsection{Dataset and Preprocessing}
We collect a real-world dataset from National Cheng Kung University Hospital (NCKUH)\footnote{The study protocol was approved by the Institutional Review Board (IRB) of NCKUH (No. B-BR-106-044 \& No. A-ER-109-027).}, containing information from 58 ICU patients. The dataset includes (i) heart rate data with 1-minute granularity from wearable devices and (ii) Apache II scores at specific time steps, used to define deterioration labels (Sec.\ref{sec:preliminary}, Def. 2) for patients. As clinical deterioration signs can appear 6–8 hours in advance~\cite{Rose2015UtilizationOE}, we extract the 8-hour heart rate time series preceding the time step when the deterioration label is assigned (the illness level increase or decreases). Table ~\ref{table:dataset} presents the preprocessed time series statistics. We randomly utilize two-thirds of the data for training and the remaining one-third for testing, reporting results as the average across 3-fold cross-validation.

\subsubsection{Baseline Methods}
We compare our \emph{TARL} with the following baselines:

\noindent\underline{Feature-based Method:} Extracts statistical features (e.g., mean, skew, standard deviation, and kurtosis) from time series as input. We include the \textbf{XGBoost} model~\cite{Chen2016XGBoostAS}

\noindent\underline{Sequence-based Method:} Classifies time series based on specific subsequences, including the classic early classification model \textbf{ECTS}~\cite{Xing2012EarlyCO}, and Shapelet Transformation (\textbf{ST})~\cite{Lines2012AST}.

\noindent\underline{Graph-based Method:} Uses graph embedding to represent time series. We compare with \textbf{Time2Graph}~\cite{Cheng2019Time2GraphRT}, which models shapelets as graph nodes and applies embedding techniques. Unlike our method, Time2Graph does not consider transition types between shapelets, nor does it preserve neighborhood and relation proximities or incorporate shapelet confidence.

\noindent\underline{\emph{TARL} Variants Method:} Evaluates the performance of \emph{TARL} with certain components removed, including the attentive mechanism (A.M.), transition confidence (T.C.), or both. Specifically, we assess \textbf{\emph{TARL}} \textbf{w/o A.M.}, \textbf{\emph{TARL}} \textbf{w/o T.C.}, and \textbf{\emph{TARL}} \textbf{w/o T.C. and A.M.}.

\begin{table}[t]
\label{tb:statistics}
\footnotesize
\centering
\renewcommand{\arraystretch}{1.0}
\begin{tabular}{|c|c||c|c|}
\hline
\multicolumn{2}{|c||}{\textbf{The Num. of TS}} & \multicolumn{2}{c|}{\textbf{Statistics of each TS}} \\ \hline\hline
\#Total & 142 & Length & 8 hr.  \\ \hline
\#Deterioration & 79  & Data granularity & 1 min.  \\ \hline
\#Recovery & 63 & Avg. missing rate & 20\% \\ \hline
\end{tabular}
\renewcommand{\arraystretch}{1}
\caption{Statistics of the preprocessed time series (TS) obtained from selected 58 ICU patients.}
\label{table:dataset}
\end{table}

\subsubsection{Evaluation Metrics}
\noindent\underline{Detection Effectiveness:} Measured using accuracy \textbf{(Acc.)}, precision \textbf{(Prec.)}, recall \textbf{(Rec.)}, F1-score \textbf{(F1)}, F0.5-score \textbf{(F0.5)}, and F2-score \textbf{(F2)}.

\noindent\underline{Detection Earliness:}  Estimates the earliness score of a detection as $\frac{\text{T}-t}{\text{T}}$, following~\cite{gupta2020approaches}, where $\text{T}$ is the total length of a time series, defined as 480 minutes (8 hours), and $t$ is the length (in minutes) of observed time series used for detecting deterioration. A higher earliness score indicates earlier detection with fewer heart rate observations. Reports include the first quartile (\textbf{Q1}), third quartile (\textbf{Q}3), interquartile range (\textbf{IQR}), and average (\textbf{Avg.}) of the earliness scores across all testing time series.

\noindent\underline{Balance Performance:} Evaluates the trade-off between detection effectiveness and earliness. We introduce a new metric, Effectiveness-Earliness Score (\textbf{EE Score}), calculated as the average of the F1-score and the average earliness score.

\begin{table*}[!t]
\small
\centering
\renewcommand{\arraystretch}{0.9}
\begin{tabular}{l|cccccc|cccc|c}
\toprule[1.3pt]
\multirow{2}{*}{\textbf{Method}} & \multicolumn{6}{c}{\textbf{Effectiveness}} & \multicolumn{4}{c}{\textbf{Earliness}} & \multicolumn{1}{c}{\textbf{Balance}} \\ \cmidrule(rl){2-7} \cmidrule(rl){8-11} \cmidrule(rl){12-12}
 & Acc.($\uparrow$) & Prec.($\uparrow$) & Rec.($\uparrow$) & F1($\uparrow$) & F0.5($\uparrow$) & F2($\uparrow$) & Q1($\uparrow$) & Q3($\uparrow$) & IQR($\downarrow$) & Avg.($\uparrow$) & EE Score ($\uparrow$)\\ 
 \midrule[1.3pt]
XGBoost & 0.53 & 0.59 & 0.76 & 0.66 & 0.62 & 0.72 & \cellcolor[gray]{0.82}{0.75} & \cellcolor[gray]{0.82}{0.95} & 0.20 & 0.73 & 0.69 \\
ECTS & 0.57 & \underline{0.67} & 0.64 & 0.65 & 0.66 & 0.65 & 0.02 & 0.13 & \cellcolor[gray]{0.82}{0.11} & 0.08 & 0.36\\
ST & 0.51 & 0.54 & \underline{0.90} & 0.68 & 0.59 & 0.79 & \cellcolor[gray]{0.82}{0.87} & \cellcolor[gray]{0.82}{0.93} & \cellcolor[gray]{0.82}{0.06} & \cellcolor[gray]{0.82}{0.83} & \underline{0.75} \\
Time2Graph & \underline{0.57} & \textbf{0.75} & 0.48 & 0.59 & \textbf{0.68} & 0.52 & 0.11 & 0.48 & 0.37 & 0.29 & 0.44 \\
\midrule
\emph{TARL} w/o T.C. & 0.55 & 0.63 & 0.81 & 0.71 & 0.66 & 0.76 & 0.50 & 0.87 & 0.37 & 0.69  & 0.70\\
\emph{TARL} w/o A.M. & 0.55 & 0.61 & 0.88 & \underline{0.72} & 0.65 & \underline{0.81} & 0.68 & 0.87 & 0.18 & \cellcolor[gray]{0.82}{0.76} & 0.74\\
\emph{TARL} w/o T.C. and A.M. & 0.52 & 0.59 & 0.69 & 0.64 & 0.61 & 0.67 & 0.62 & 0.80 & 0.17 & 0.70 & 0.67 \\
\midrule
\emph{\textbf{TARL}} \textit{(ours)} & \textbf{0.61} & 0.62 & \textbf{0.92} & \textbf{0.74} & \underline{0.67} & \textbf{0.84} & \cellcolor[gray]{0.82}{0.72} & \cellcolor[gray]{0.82}{0.88} & \cellcolor[gray]{0.82}{0.16} & \cellcolor[gray]{0.82}{0.79} & \textbf{0.76}\\ 
\bottomrule[1.3pt]
\end{tabular}
\caption{Performance comparison. The best and second-best results are in bold and underlined, respectively. Cells in gray indicate that the earliness performance ranks among the top three across all methods.}
\label{table:rq1_effect}
\end{table*}

\subsubsection{Implementation Details}
For training \emph{TARL}, the embedding dimension is set to 256, with 20 shapelets of length 15. The weight $\omega$ (Eq.~\ref{eq:head_update}) is set to 0.5. 
XGBoost serves as the classifier for \emph{TARL} in our experiments. To simulate wearable device monitoring, testing time series are split into 30-minute segments and sent sequentially as real-time situation. Experiments are conducted on a system with 12 CPU cores and 64GB RAM, running CUDA 12.1.

\subsection{Performance Comparison Results}

The comparative results are shown in Table~\ref{table:rq1_effect}. The key observations are drawn as follows:

\begin{itemize}[leftmargin=*]
    \item \textbf{Effectiveness.} \emph{TARL} achieves the highest accuracy, recall, F1-score, and F2-score. While Time2Graph outperforms \emph{TARL} in precision and F0.5-score, its significantly lower recall indicates it misses many deterioration cases. In addition, \emph{TARL} achieves the highest accuracy, ensuring reliable overall classification. In early detection of patient deterioration, prioritizing recall is essential to identify as many at-risk patients as possible, even at the cost of some false positives. Since F1-score and F2-score emphasize recall more than F0.5-score, \emph{TARL} provides a better detection effectiveness, making it more practical for real-world use.
    \item \textbf{Earliness and Balance.} While \emph{TARL} isn't the top performer in earliness, it consistently ranks in the top three and achieves the highest Effectiveness-Earliness (EE) score, demonstrating a strong balance between reliability and early detection. Notably, when \emph{TARL} attains the highest accuracy, recall, F1-score, and F2-score, its average earliness score reaches 0.79, meaning it detects deterioration about 379 minutes (approximately 6 hours) in advance while observing only 30\% of a patient’s heart rate time series. This capability is particularly significant for \textit{myocardial infarction (MI)}, a common ICU condition, where the golden window for treatment is about 6 hours~\cite{Steinberg1994EffectsOT}. These results highlight \emph{TARL} as the most reliable for detecting deterioration within the golden time of treatment, demonstrating its high practicability for real-world applications.
    \item \textbf{Ablation Study.} Removing both transition confidence and the attentive mechanism (\emph{TARL} w/o T.C. and A.M.) results in the worst detection effectiveness and the lowest EE score. Excluding only transition confidence (\emph{TARL} w/o T.C.) causes a larger drop in earliness score, likely because the impact of missing values is ignored during embedding, weakening reliable shapelet relationships and delaying deterioration detection. On the other hand, removing the attentive mechanism (\emph{TARL} w/o A.M.) causes a 6\% drop in accuracy. This may be because structurally similar shapelets often exhibit similar future changes, and ignoring these relationships may miss meaningful heart rate transitions. These findings confirm the importance of both transition confidence and attentive mechanism in \emph{TARL}.
\end{itemize}

\begin{figure}
\graphicspath{{figs/}}
\begin{center}
\includegraphics[width=0.48\textwidth]{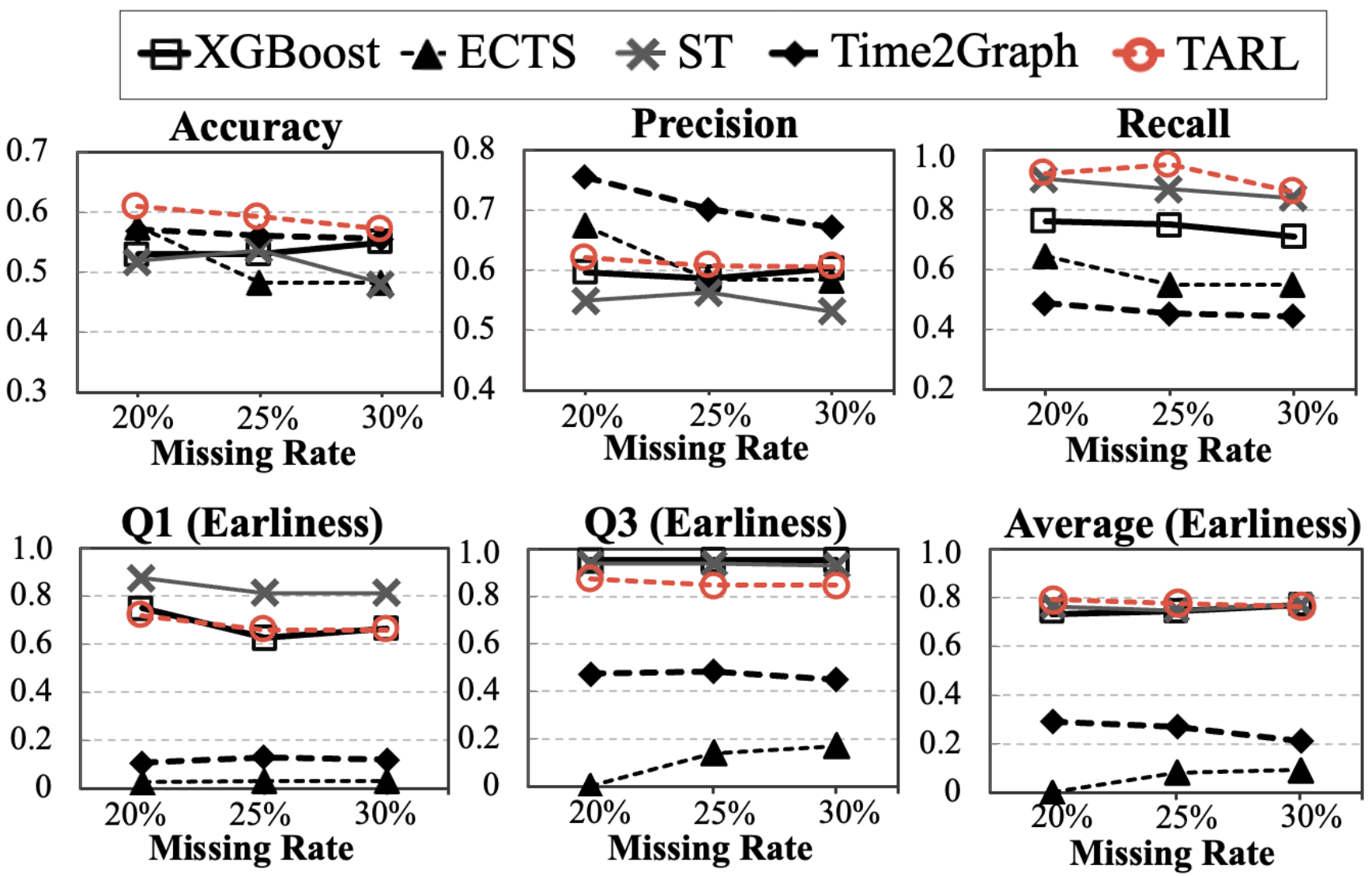}
\end{center}
\caption{Performance of \emph{TARL} with different missing rates.}
\label{fig:rq2_missing_rate}
\end{figure}

\subsection{Performance on Different Data Missing Rates}

To simulate missing data in wearable devices, we vary the missing rate of heart rate data to \{20\%, 25\%, 30\%\} to evaluate \emph{TARL}'s performance. As shown in \figurename~\ref{fig:rq2_missing_rate}, \emph{TARL}'s performance varies only slightly across rates. A more detailed analysis is provided below.
\begin{itemize}[leftmargin=*]
    \item  Although \emph{TARL}'s performance slightly declines with higher missing rates, its accuracy and recall consistently surpass baselines. While Time2Graph achieves the highest precision, its precision drops significantly with more missing data, whereas \emph{TARL} maintains stable precision, demonstrating its robustness to missing values in wearable data.
    \item For earliness performance, \emph{TARL} maintains an average earliness score of approximately 0.8 across different missing rates, showing its ability to achieve competitive early detection even with varying levels of missing data. This highlights its practicality for early patient deterioration detection using wearable device data.
\end{itemize}

\subsection{Parameter Sensitivity Analysis}

\begin{figure}[ptb]
\graphicspath{{figs/}}
\subfloat[Effectiveness Performance.] {
\label{fig:rq4_omega_effect}
\includegraphics[width=0.24\textwidth]{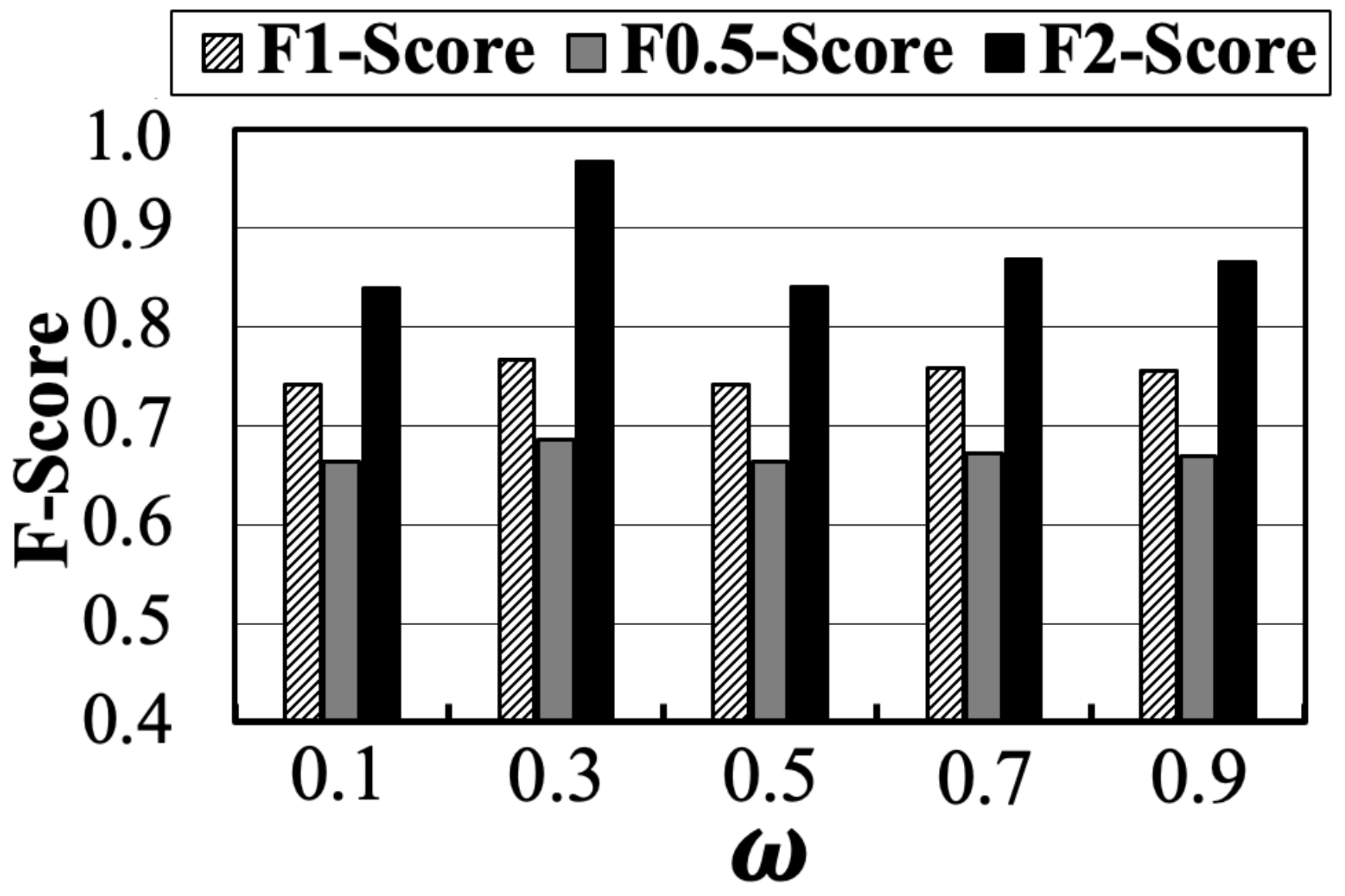}
}
\subfloat[Earliness Performance.] {
\label{fig:rq4_omega_earliness}
\includegraphics[width=0.24\textwidth]{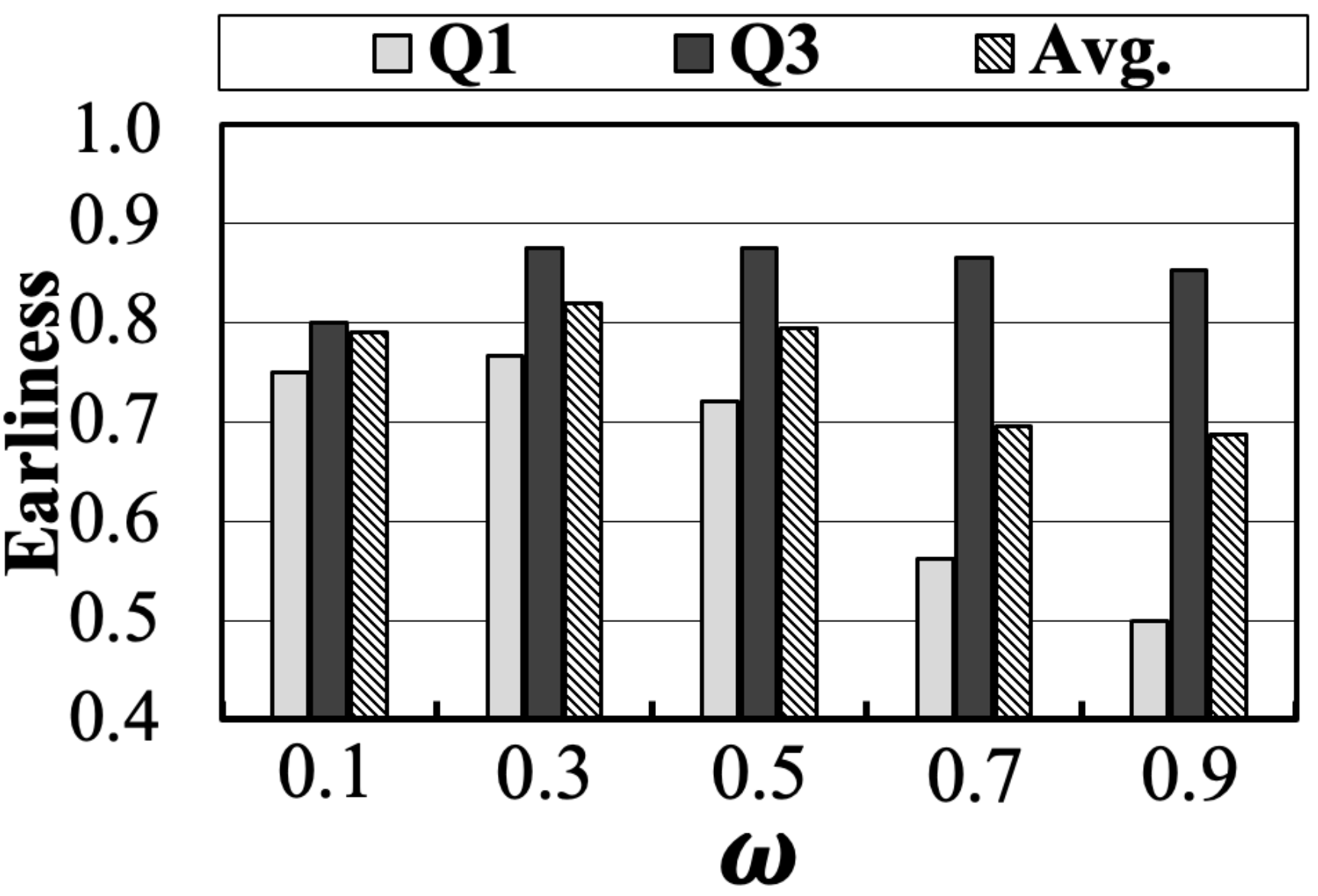}
}
\caption{The performance of \emph{TARL} with varying $\omega$.}%
\label{fig:rq4_omega}

\end{figure}

\begin{figure}[ptb]
\graphicspath{{figs/}}
    \subfloat[F1-score.]{\label{fig:rq4_dim_sh_f1}\includegraphics[width=0.24\textwidth]{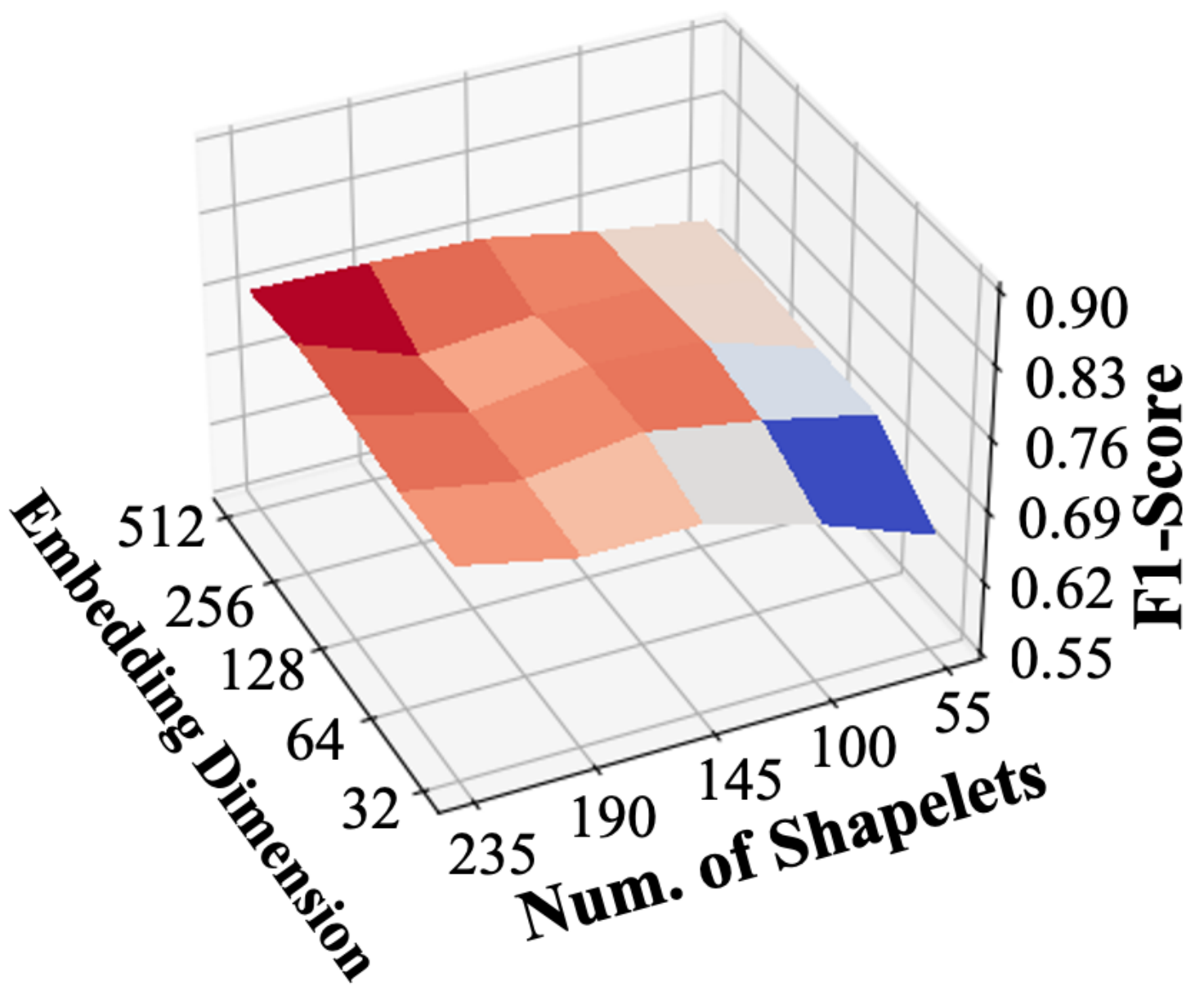}} 
    \subfloat[Average Earliness.]{\label{fig:rq4_dim_sh_early}\includegraphics[width=0.24\textwidth]{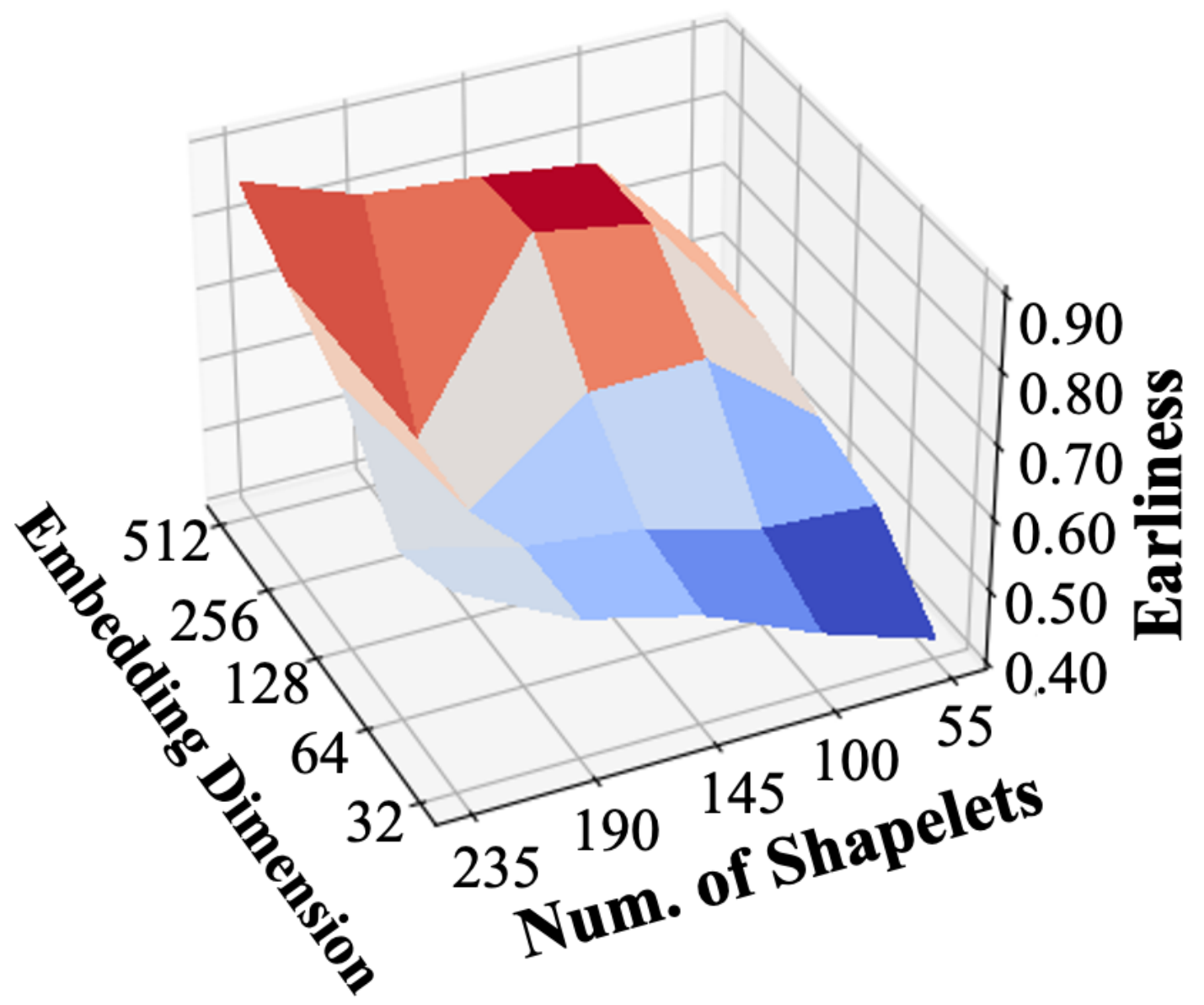}}
    \caption{The performance of \emph{TARL} with varying the embedding dimension and the number of shapelets. Within each metric, higher values are represented in red, while lower values are shown in blue.}
    \label{fig:rq4_dim_sh}
\end{figure}

To analyze parameter sensitivity in \emph{TARL}, we conduct two experiments: one on the weight ($\omega$) balancing neighborhood and relation proximities in Eq.~\ref{eq:head_update}, and the other on the embedding dimension and the number of selected key shapelets. Results are shown in \figurename~\ref{fig:rq4_omega} and \figurename~\ref{fig:rq4_dim_sh}, respectively.

\begin{itemize}[leftmargin=*]
    \item As shown in \figurename~\ref{fig:rq4_omega}, setting $\omega$ to 0.3 yields the best F-score and improved earliness. While F-scores remain relatively stable across $\omega$, earliness performance is more sensitive. As $\omega$ increases, the earliness score gap widens, possibly due to ignoring relation proximity during embedding learning, weakening shapelet transition time interval reinforcement. \emph{TARL} struggles to accurately capture when a patient's condition is likely to change, effecting the earliness of the detection.
    \item In \figurename \ref{fig:rq4_dim_sh}, the performance of \emph{TARL} is evaluated by varying the embedding dimension as \{32, 64, 128, 256, 512\} and the number of selected key shapelets as \{55, 100, 145, 190, 235\}. While F-scores gradually improve as both embedding dimension and the number of shapelets increase, they drop sharply when fewer than 145 shapelets are used. This is likely because a smaller shapelet set results in an insufficiently structured shapelet-transition KG. For earliness performance, the embedding dimension shows higher sensitivity, possibly because higher-dimensional embeddings better preserve time interval information in shapelet transitions, improving the detection earliness.
\end{itemize}

\begin{figure}[t]
\graphicspath{{figs/}}
\begin{center}
\includegraphics[width=0.46\textwidth]{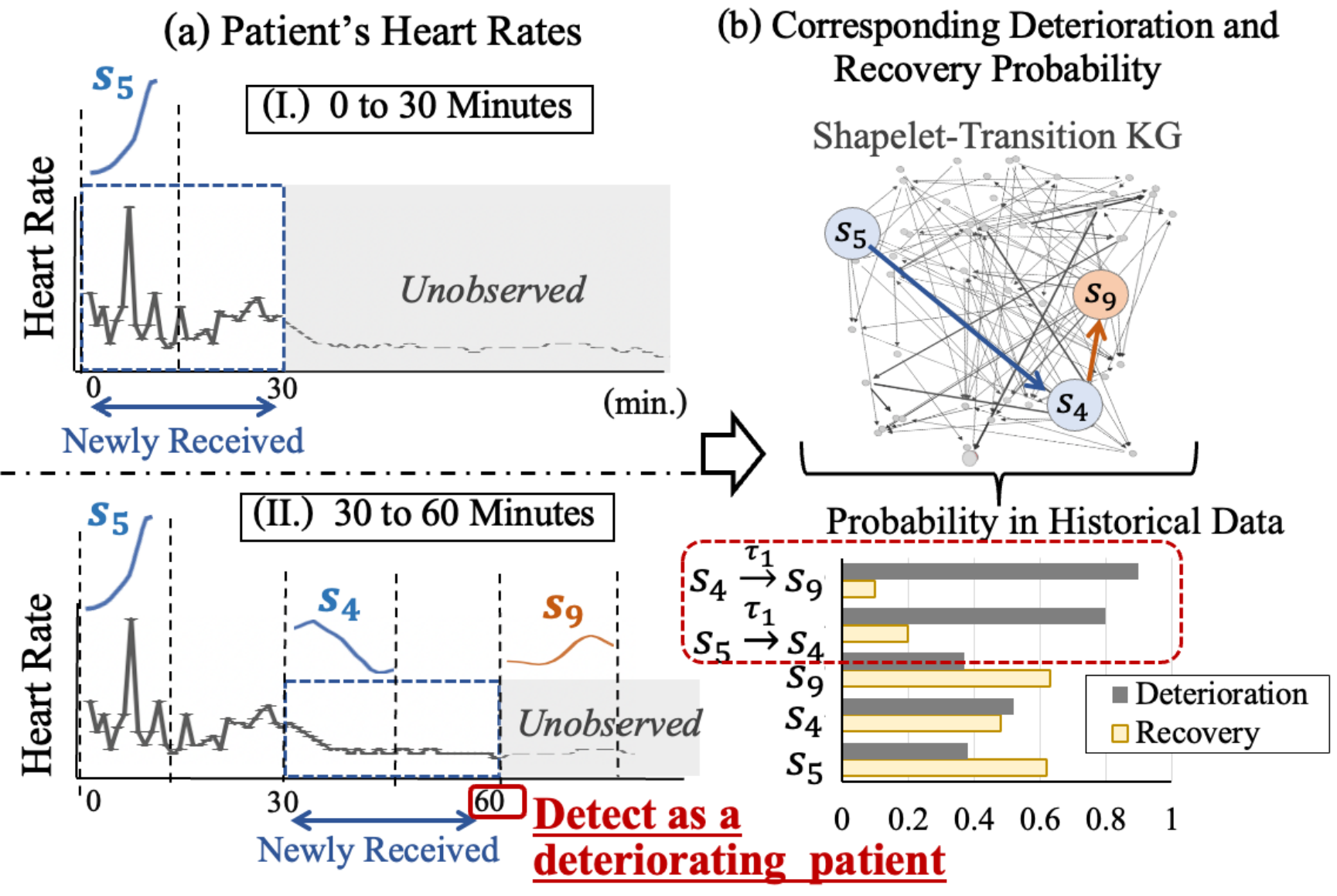}
\end{center}
\caption{Visualization of the detection process for a deteriorating patient in two detection windows. (a) displays the newly received heart rates (observed information) and the matched shapelets; (b) depicts corresponding probabilities of observed shapelet relationships in recovering and deteriorating patients from historical data.}
\label{fig:case_study}
\end{figure}

\subsection{Case Study}
We present a case study in \figurename~\ref{fig:case_study}, illustrating the detection process for a patient successfully identified as deteriorating with 60 minutes of heart rate monitoring.

\noindent\underline{Time window (0 to 30 minutes)}: Shapelet $s_5$ is identified, but $s_5$ has a low deterioration probability. Therefore \emph{TARL} does not detect it at this stage.

\noindent\underline{Time window (30 to 60 minutes)}: Shapelet $s_4$ is identified. While $s_4$ alone still has a low deterioration probability, its transition from $s_5$ through relation $\tau_1$ (transition within 0-30 minutes) exhibits a high deterioration probability. Moreover, $s_9$ is identified as having a transition relation following $s_4$ in shapelet-transition KG, and it also appears in the patient's heart rates in the next 30 minutes. The potential transition from $s_4$ to $s_9$ through relation $\tau_1$ further reinforces the high deterioration probability, leading \emph{TARL} to detect signs of deterioration within this time window. This analysis shows that modeling shapelet relationships captures early deterioration signs and potential future transitions. It demonstrates that \emph{TARL} delivers accurate and explainable detection.
\section{Conclusion}

In this paper, we introduce a novel representation framework, \emph{TARL}, designed for real-time monitoring using wearable devices to enable early detection of illness deterioration by learning patient heart rate representations. We create a shapelet-transition knowledge graph to model heart rate transitions and devise a transition-aware embedding learning method to enhance shapelet relationships and assess missing data impact.
With help from physicians and nurses, we gather ICU patient heart rate data from wearables and diagnostic metrics as deterioration indicators. Experimental results show that \emph{TARL} achieves the best balance between detection effectiveness and earliness, maintaining strong performance even with varying levels of missing data. A case study also illustrates \emph{TARL}’s explainable detection process, which has the potential to help clinicians identify early signs of patient deterioration.

\section*{Acknowledgments}

We sincerely thank Prof. Nai-Ying Ko, Dr. Yi-Ting Chung, and Dr. Chang-Chun Chen from the Department of Nursing, and Prof. Yu-Chen Shu from the Department of Mathematics at National Cheng Kung University (NCKU), for their invaluable support in wearable data collection, patient coordination, and access to physiological data from NCKU Hospital.

In addition, this paper was supported in part by National Science and Technology Council (NSTC), R.O.C., under Contract 113-2221-E-006-203-MY2, 114-2622-8-006-002-TD1 and 113-2634-F-006-001-MBK.

\bibliographystyle{named}
\bibliography{bibliography}

\end{document}